\pdfoutput=1

\documentclass[11pt]{article}

\usepackage[final]{acl} 

\usepackage{times}
\usepackage{comment}
\usepackage{latexsym}
\usepackage{afterpage}
\usepackage{hyperref}
\usepackage{svg}
\usepackage[T1]{fontenc}

\usepackage[utf8]{inputenc}

\usepackage{microtype}

\usepackage{inconsolata}
\usepackage{multirow}
\usepackage{longtable}
\usepackage{array}
\usepackage{graphicx}

\title{A comprehensive survey of contemporary Arabic sentiment analysis: Methods, Challenges, and Future Directions} 

\author{Zhiqiang Shi \\
   University of Edinburgh\\
  \texttt{shizhiqiang126@163.com} \\\And
  Ruchit Agrawal \\
  University of Birmingham 
  \\ 
  \texttt{r.r.agrawal@bham.ac.uk} \\} 

\begin{document}
\maketitle
\begin{abstract}
\vspace{-0.1cm}
Sentiment Analysis, a popular subtask of Natural Language Processing, employs computational methods to extract sentiment, opinions, and other subjective aspects from linguistic data. Given its crucial role in understanding human sentiment, research in sentiment analysis has witnessed significant growth in the recent years. However, the majority of approaches are aimed at the English language, and research towards Arabic sentiment analysis remains relatively unexplored. This paper presents a comprehensive and contemporary survey of Arabic Sentiment Analysis, identifies the challenges and limitations of existing literature in this field and presents avenues for future research. 
We present a systematic review of Arabic sentiment analysis methods, focusing specifically on research utilizing deep learning. We then situate Arabic Sentiment Analysis within the broader context, highlighting research gaps in Arabic sentiment analysis as compared to general sentiment analysis. Finally, we outline the main challenges and promising future directions for research in Arabic sentiment analysis. 
\end{abstract}
\section{Introduction}
\vspace{-0.1cm}
Sentiment Analysis (SA), also referred to as opinion mining, leverages computational models to extract individuals' sentiments and opinions from data \cite{sentiment-analysis-book}. This field has garnered significant attention from both academic and industrial sectors, as evidenced by the multitude of studies conducted to comprehend human sentiment \cite{sentiment-analysis-book}, \cite{zeng-li-2022-survey}, \cite{https://doi.org/10.1002/asi.23716}. However, although Arabic is a widely popular language spoken by over 372 million people across the globe, the volume of research dedicated to Arabic Sentiment Analysis (ASA) remains considerably lower compared to high-resourced languages such as English and French. 
This study presents a systematic review of existing literature on Arabic sentiment analysis, with a particular focus on approaches that employ deep learning methodologies. 
\par Several prior surveys on Arabic sentiment analysis (ASA) exist \cite{Almurqren2024ArabicTS}, \cite{ALAYYOUB2019320}, \cite{arabic_survey_subjectivity}, \cite{arabic_survey_sc}, \cite{arabic_survey_asa_social_media}, \cite{arabic_survey_asa}, \cite{arabic_survey_sarcasm_detection}. However, the majority of these do not cover contemporary deep learning based methods. Additionally, 
these do not present a detailed analysis of the gaps, challenges and future directions for ASA.  
This paper fills this gap and presents a comprehensive survey of contemporary methods for Arabic sentiment analysis, systematically organizing recent literature in the field and highlighting the key contributions and limitations of current SA methods. We also situate these approaches within the broader framework of general sentiment analysis and approaches for high-resource languages, facilitating an understanding of the developments as well as the gaps in ASA. The major contributions of this study are summarised below:

\begin{itemize}
\vspace{-0.2cm}
\item We present a systematic survey of contemporary research conducted in Arabic sentiment analysis, with a focus on deep learning methodologies. We present an analysis of the key contributions and limitations of state-of-the-art ASA methods across various dimensions such as modality (uni-modal, multi-modal), granularity (coarse-grained, fine-grained) and context (sentence-level, document-level, aspect-level). 
\vspace{-0.2cm}
\item We situate Arabic sentiment analysis within the broader framework of general sentiment analysis, identifying research gaps in Arabic sentiment analysis, and highlighting areas where advancements are needed to bridge the gap with high-resource languages.
\vspace{-0.2cm}
\item We outline the key challenges in developing robust sentiment analysis models for the Arabic language, and present promising directions to guide future research in this field. 
\end{itemize}

\section{The Evolution of Arabic Sentiment Analysis} 
In this section, we describe the evolution of Arabic sentiment analysis, from lexicon based methods to deep learning based methods. To highlight the importance of these traditional methods, we conduct a case study using Arabic lexicons, in which we highlight how lexicons can be utilised to improve deep learning based methods. 
\vspace{-0.2cm}
\subsection{Lexicon Based Methods} \vspace{-0.1cm}
Lexicon based methods utilise a pre-defined lexicon to determine the sentiment of the given text. The words in the lexicon are annotated with polarity or sentiment scores. The overall sentiment of the text are calculated by summing up all the words' sentiment score. Given their crucial role in lexicon based methods, we briefly mention some widely used Arabic lexicons. 

\textbf{Arabic Senti-Lexicon} \cite{senti-lexicon}: Arabic Senti-Lexcion contains 3880 synsets that are annotated with part of speech, polarity scores and inflected forms. 

\textbf{ArsenL (Arabic Sentiment Lexicon)} \cite{badaro-etal-2014-large}: ArsenL are constructed from multiple resources, including English WordNet (EWN), Arabic WordNet (AWN), English SentiWordNet (ESWN), and SAMA (Standard Arabic Morphological Analyzer). It contains 157969 synsets and has positve, negative and neutral sentiment scores.  

\subsection{Machine learning based methods}
Lexicon based methods are simple and fast, but they heavily rely on the lexicons and the sentiment scores of the lexicons lack context. Machine learning based methods can help to overcome these limitations by learning the sentiment features from data rather than pre-defining them by humans. However, for traditional machine learning methods, feature engineering is required as a precursor to the ML algorithms. \par Some widely used feature engineering methods include bag-of-words \cite{bag-of-words}, TF-IDF \cite{tf-idf} and word embedding \cite{Almeida2019WordEA}. After the features have been extracted, machine learning methods such as naive Bayes \cite{AlHoraibi2016SentimentAO}, support vector machines \cite{svm_sa} and random forests based methods \cite{Altawaier2016ComparisonOM} can be used for sentiment analysis. In the following sections, we will systematically review deep learning based approaches.

\subsection{The importance of traditional methods: a case study on sentiment lexicons} 
It is important to note that even with the rise of deep learning based methods for sentiment analysis, sentiment lexicons like ArSenL\cite{badaro-etal-2014-large} can still be valuable as they provide a foundational understanding of sentiment that can enhance model performance, especially in low-resource scenarios or when dealing with domain-specific language that may not be well-represented in training data. Some use cases of these lexicons include data preprocessing where the irrelevant terms are filtered out based on the sentiment lexicon, sentiment weighting \cite{Zhang2011CombiningLA} where the lexicon is employed to help the model weight sentiment-relevant features more effectively. In the following paragraphs, we will use some examples to illustrate how sentiment lexicons can be utilised to improve the sentiment analysis performance. 

\textbf{Feature Augmentation}: Lexicons can be utilised to augment the features. In \cite{Heikal2018SentimentAO}, a sentiment lexicon is integrated to augment the features for deep learning based modes. \cite{Xiang2021LexicalDA} explores part-of-speech-focused lexical substitution for data augmentation to enhance sentiment analysis performance. \cite{lexicon_augmentation} discusses how lexicon-based approaches assign sentiment polarities and scores to keywords, which can be used for feature augmentation in hybrid models. 

\textbf{Interpretability}: Arabic sentiment lexicons enhance interpretability in sentiment analysis by providing a clear framework for understanding how specific words and phrases contribute to sentiment assessments. By combining lexicons with advanced methods like attention-based LSTM and explainable AI techniques, such as LIME \cite{Ribeiro2016WhySI}, researchers can further clarify which features are most significant in determining sentiment, thereby enhancing the transparency of their findings \cite{Abdelwahab2022JustifyingAT}. These lexicons facilitate the identification of the sentiment polarity of individual terms, making it easier to trace the reasoning behind a sentiment classification \cite{lexical_interpretability}. Moreover, the integration of lexicon-based approaches with machine learning techniques can improve the interpretability of complex models, as researchers can analyze how lexicon entries influence the overall sentiment predictions \cite{Ambreen2024PredictingCS}.  

\begin{table*} 
  \centering
  \begin{tabular}{m{5cm} m{1.2cm} m{2cm} m{2.7cm} m{3cm}}
    \hline
    \textbf{Dataset} & \textbf{Modality} & \textbf{Granularity} & \textbf{Context} & \textbf{Dialect} \\
    \hline
    LABR \cite{aly-atiya-2013-labr} & text & SC & document-level & MSA  and various other dialects \\ 
    ASTD \cite{nabil-etal-2015-astd} & text & SC & document-level & MSA and Egyptian Arabic \\ 
    ArSentD-LEV \cite{DBLP:journals/corr/abs-1906-01830} & text & SC, ABSA, MAST & document-level & Levantine dialect \\ 
    ArSarcasm \cite{abu-farha-magdy-2020-arabic} & text & SC, MAST & document-level & various dialects\\ 
    ArSarcasm-v2 \cite{abu-farha-etal-2021-overview} & text & SC, MAST & document-level & various dialects \\
    Arsen-20 \cite{fang2024arsen} & text & SC & document-level & various dialects \\  
    Arabic multimodal dataset \cite{Haouhat2023TowardsAM} & text, audio, video & SC & document-level (video segments) & various dialects \\ 
    \hline
  \end{tabular}
  \captionsetup{justification=centering}
  \caption{\label{table: datasets} Datasets for Arabic Sentiment Analysis, organised according to modality, granularity and context.\\ \textit{SC: Sentiment Classification \\ MAST: Multifacted Analysis of Subjective Text \\ ABSA: Aspect based Sentiment Analysis.}}
\end{table*}
\begin{table*} 
\centering 
  \begin{center}
  \begin{tabular}[h]{m{2cm} m{6cm} m{5.5cm} m{1.5cm}}
  \hline 
   \textbf{Methods} & \textbf{Contributions} &\textbf{Limitations} & \textbf{Accuracy} \\
   \hline 
  \cite{dahou-etal-2016-word} & Develops Arabic word embeddings and employs CNN as a classifier & Task-specific method, static word embeddings & LABR (89.6\%), ASTD (79.07\%) \\ 
  \cite{medhaffar-etal-2017-sentiment} & Annotates a Tunisian dialect corpus and evaluates models on different dialects & Only experiments with traditional machine learning methods & -  \\ 
  \cite{baly-etal-2017-characterization} & Performs a characterization study that analyses tweets from different Arab regions, and compares ML-based vs. deep-learning methods for Arabic SA & Does not experiment on different dialects and topics & ASTD (SVM 51.7\%, RNTN 58.5\%)  \\ 
  \cite{Guellil2018SentiALGAC} & Automatically constructs an Algerian dialect corpus & Evaluation is carried out only using traditional ML methods & -  \\  \\
  \cite{attia-etal-2018-multilingual} & Proposes a language independent, multi-class model for SA using CNNs & Evaluation for Arabic is only carried out on ASTD \cite{nabil-etal-2015-astd}, unclear whether the model generalises well to other datasets & ASTD (67.93\%)  \\  \\
  \cite{Alyafeai2021EvaluatingVT} & Compares different tokenizers for different Arabic classification tasks & Does not evaluate on complex architecture like attention-based models -  \\ \\
  \cite{Atabuzzaman2023ArabicSA} & Proposes an explainable sentiment classification framework for Arabic & Does not conduct experimentation on Transformer-based models & LABR (88.0\%) \\ 
  \hline  
  \end{tabular}
  \end{center}
\captionsetup{justification=centering}
  \caption{\label{tabel: task-specific sc} Task-specific methods for Arabic sentiment classification. \\ \textit{LABR: Large Scale Arabic Book Reviews Dataset \\ ASTD: Arabic Sentiment Tweets Dataset}}
\end{table*}

\section{Situating ASA methods vis-à-vis general SA approaches}

We situate the research in Arabic Sentiment Analysis (ASA) and juxtapose it with general trends in sentiment analysis (SA) in this section. We present an overview of several sentiment analysis tasks, and for each task we highlight the advancements in  general sentiment analysis research, followed by a focus on Arabic-specific sentiment analysis. While not exhaustive, the selected approaches illustrate key differences and current trends between general and Arabic sentiment analysis.
Table \ref{table: datasets} provides an overview of datasets for ASA organised according to the modality, granularity and context involved. We refer to these datasets in the subsequent subsections. 

\vspace{-0.3cm}
\subsection{Sentiment Classification}
Sentiment classification involves assigning a sentiment label (positive, neutral, negative) or a sentiment rating (e.g., one to five) to a given input, which can be text or data from other modalities. As one of the earliest sentiment analysis tasks \cite{sentiment-analysis-book}, sentiment classification has attracted significant research interest.
\subsubsection{General Sentiment Classification} \label{section: general sc}

The development of general sentiment classification reflects the ongoing paradigm shifts within the field of natural language processing \cite{Liu2021PretrainPA}. Early works in sentiment classification primarily relied on task-specific models, employing either traditional machine learning methods like Support Vector Machines (SVMs) or deep learning-based approaches. These models were trained on labeled data and limited to solving specific tasks. 

\par However, the emergence of pre-trained language models such as BERT \cite{devlin-etal-2019-bert} has revolutionized the field. These models, typically based on components of the Transformer architecture, are pre-trained on massive amounts of unlabeled data and subsequently fine-tuned for specific tasks, including sentiment classification. Large language models like GPT-3 \cite{Brown2020LanguageMA} further push the boundaries of model size, demonstrating the ability to acquire various emergent capabilities such as in-context learning when scaled sufficiently large \cite{Wei2022EmergentAO}. A systematic analysis of large language models' effectiveness in tackling various sentiment analysis tasks, including sentiment classification, is provided by \cite{Zhang2023SentimentAI}.
\vspace{-0.3cm}
\subsubsection{Arabic Sentiment Classification} \label{section: arabic sentiment classification}
The development of Arabic sentiment classification follows a similar trajectory to that of general sentiment classification. Early research predominantly focused on task-specific models trained on sentiment classification datasets for Arabic text. \cite{dahou-etal-2016-word} constructed Arabic word embeddings and subsequently employed a Convolutional Neural Network (CNN) as a classifier. \cite{attia-etal-2018-multilingual} proposed a language-independent framework for text classification, evaluating its performance on Arabic sentiment classification tasks as well. Table \ref{tabel: task-specific sc} provides detailed descriptions of various task-specific methods for Arabic sentiment classification, along with their contributions and limitations. As highlighted in the table, the biggest limitation of such methods is that they are task-specific and do not generalise well to other tasks or dialects. 

\par With the remarkable success of pre-trained language models based on bidirectional transformers, such as BERT \cite{devlin-etal-2019-bert}, on diverse natural language understanding tasks, numerous studies have explored their utilization for Arabic sentiment classification. \cite{eljundi-etal-2019-hulmona} developed hULMonA, a pre-trained language model specifically for Arabic, and fine-tuned it for Arabic sentiment analysis. AraBERT \cite{antoun-etal-2020-arabert} builds upon this work by pre-training the model entirely on an Arabic corpus and evaluating its performance on various tasks. \cite{abdul-mageed-etal-2021-arbert} introduced ARBERT and MARBERT, language models pre-trained on dialectal Arabic. Table \ref{tabel: language model sc} offers detailed descriptions of different pre-trained language model-based methods for Arabic sentiment classification.
\begin{table*} 
\centering 
  \begin{center}
  \begin{tabular}{m{2cm} m{5cm} m{4.5cm} m{3cm}}
  \hline 
  \textbf{Methods} & \textbf{Contributions} &\textbf{Limitations} & \textbf{Accuracy} \\
  \hline 
   hULMonA \cite{eljundi-etal-2019-hulmona} & Develops a pre-trained LM for Arabic and fine-tunes it for SA & Does not employ an Arabic specific tokenizer and only evaluates on the SA task & ASTD (86.5\%), ArSenTD-LEV (52.4\%) \\ \\
  AraBERT \cite{antoun-etal-2020-arabert} & Pre-trains an Arabic LM AraBERT and evaluates performance on different tasks & Does not systematically evaluate the model on different dialects & LABR (86.7\%), ASTD (92.6\%), ArSenTD-Lev (59.4\%) \\ \\
  \cite{alkaoud-syed-2020-importance} & Proposes tokenization methods for static and contextual word embeddings and improves their performance & Does not study generalisation ability of the proposed method & LABR (89.87\%) \\ \\
  \cite{abdul-mageed-etal-2021-arbert} & Introduces ARBERT and MARBERT, pre-trains models on dialectal Arabic, introduces ARLUE benchmark & The models have a high memory requirement, thereby impeding computational efficiency & LABR (92.51\%), ASTD (95.24\%), ArSenTD-Lev (61.38\%) \\ \\
  \cite{alyafeai-ahmad-2021-arabic} & Employs distillation and quantization to train compact Arabic language models & The effects of hyperparameter tuning are not analysed & LABR (87.5\%), ASTD (86.2\%)  \\ \\
  \cite{el-mekki-etal-2021-domain} & Introduces an unsupervised domain adaptation method for Arabic cross-domain and cross-dialect SA & Does not study the effect of domain adaptation from high-resource languages to Arabic & -  \\ \\
  \cite{abu-kwaik-etal-2022-pre} & Compares feature-based, deep learning and pre-trained LM based methods on dialectal Arabic SA & Lacks an error analysis and a comparison of feature-based vs pre-trained LMs in different situations & -  \\ \\
  \cite{Refai2022DataAU} & Proposes a data augmentation method for Arabic text classification using Transformer based models & Does not study the generalisation ability of their method & -  \\ \\
  \hline 
  \end{tabular}
  \end{center}
  \captionsetup{justification=centering}
  \caption{\label{tabel: language model sc} Pre-trained language model (LM) based methods for Arabic sentiment classification. \\ \textit{hULMonA: The First Universal Language Model in Arabic \\ LABR: Large Scale Arabic Book Reviews Dataset \\ ASTD: Arabic Sentiment Tweets Dataset \\
  ArSenTD-Lev: Arabic Sentiment Twitter Dataset for the Levantine dialect}}
\end{table*}

\subsection{Multifaceted Analysis of Subjective Text (MAST)}

Multifaceted analysis of subjective text (MAST) represents an extension of sentiment classification that delves deeper into task granularity. It shifts the focus towards more specialized tasks, such as irony detection \cite{zeng-li-2022-survey} and comparative opinion mining \cite{https://doi.org/10.1002/asi.23716}.

\subsubsection{General MAST}

The development trajectory of general MAST mirrors that of general sentiment classification, as discussed in Section \ref{section: general sc}. Due to the focus on specialized tasks within MAST, it encompasses a multitude of sub-tasks. While these sub-tasks have been extensively explored in the field, a detailed description falls outside the scope of this survey. We encourage readers to refer to comprehensive surveys on specific sub-tasks, such as those by \cite{zeng-li-2022-survey} and another work referenced here \cite{https://doi.org/10.1002/asi.23716}.

\subsubsection{Arabic MAST}

Compared to general MAST, research on Arabic MAST remains less developed. This section will solely introduce research on Arabic sarcasm detection, as it has garnered a relatively larger body of work following the release of the ArSarcasm \cite{abu-farha-magdy-2020-arabic} and ArSarcasm-v2 \cite{abu-farha-etal-2021-overview} datasets, alongside a shared task organized by WANLP \cite{abu-farha-etal-2021-overview}. 

\par While various methods have been employed, including traditional machine learning approaches, task-specific deep learning methods, and pre-trained language model-based methods, the latter category combined with various optimizations has emerged as the most effective approach. \cite{hengle-etal-2021-combining} propose a hybrid model that leverages contextual representations from AraBERT \cite{antoun-etal-2020-arabert} alongside static word vectors. 

\par Additionally, recent research explores various machine learning techniques such as down-sampling and augmentation \cite{israeli-etal-2021-idc} for this task. \cite{faraj-etal-2021-sarcasmdet} employ an ensemble approach, combining different pre-trained language models with a hard voting technique. \cite{Talafha2021SarcasmDA} propose framing the problem as a regression task, predicting the level of sarcasm. Table \ref{tabel: sarcasm detection} provides detailed descriptions of methods for Arabic sarcasm detection, along with their contributions and limitations.

\begin{table*} 
\centering 
  \begin{center} 
  \begin{tabular}[h]{m{2cm} m{6.2cm} m{4.5cm} m{2cm}}
  \hline 
  \textbf{Methods} & \textbf{Contributions} & \textbf{Limitations} & \textbf{Accuracy}  \\
  \hline 
  \cite{hengle-etal-2021-combining} & Proposes a hybrid model which combines contextual representations from AraBERT \cite{antoun-etal-2020-arabert} and static word vectors & Hybrid model, computational efficiency is impeded owing to use of both contextual and static embeddings. & 74.1\%  \\ \\
  \cite{israeli-etal-2021-idc} & Employs pre-trained Transformer based models and various machine learning techniques such as down-sampling and augmentation & Does not explain the effects of these techniques & 76.7\%  \\ \\
  \cite{Talafha2021SarcasmDA} & Annotates an Arabic sarcasm detection dataset, trains a regression model and predicts the level of sarcasm & While the model can output the level of sarcasm, it is reliant on a binary classification dataset & -  \\ \\
  \cite{khondaker-etal-2022-benchmark} & Applies contrastive learning to Arabic social meaning tasks & Does not study the generalisation ability of their method & -  \\ \\
  \cite{faraj-etal-2021-sarcasmdet} & Ensembles different pre-trained language models and employs hard voting technique & The method is not efficient as it needs various pre-trained language models. & 78.3\%  \\ \\
  \cite{el-mahdaouy-etal-2021-deep} & Proposes an end-to-end multi-task model for Arabic sentiment analysis and sarcasm detection & Does not present experimentation on other tasks & 76.8\%  \\ \\
  \cite{kaseb-farouk-2022-saids} & Proposes SAIDS that uses its prediction of sarcasm and dialect as known information to predict the sentiment & Does not study the generalisation ability of their method & -  \\ 
  \hline 
  \end{tabular}
  \end{center}
  \vspace{-0.2cm}
  \caption{\label{tabel: sarcasm detection} Methods for Arabic sarcasm detection. The accuracy is evaluated on ArSarcam-v2 dataset.} 
  \vspace{0.2cm}
\end{table*}

\begin{table*} 
\centering 
\begin{center}
  \begin{tabular}[h]{m{3cm} m{6cm} m{6cm}}
  \hline 
  \textbf{Methods} & \textbf{Contributions} & \textbf{Limitations} \\  
  \hline 
  \cite{inbook} & Employs pre-trained word embeddings for Arabic ABSA & Only uses traditional machine learning methods as classifier \\ 
  \cite{AlSmadi2017DeepRN} & Compares RNN and SVM for Arabic ABSA & The dataset is relatively small, does not use other deep learning models such as LSTM \\ 
  \cite{Alshammari2020AspectbasedSA} & Compares CNN and traditional machine learning methods for Arabic ABSA & Task-specific method, does not compare different deep learning methods \\ 
  \cite{AlDabet2021EnhancingAA} & Proposes different network architectures for various Arabic ABSA tasks & Task-specific method \\ 
  \cite{Abdelgwad2021ArabicAS} & Develops a BERT based model for Arabic ABSA & Does not present experimentation on other tasks \\ 
  \hline 
  \end{tabular}
  \end{center}
  \caption{\label{Table: arabic absa} Methods for Arabic aspect based sentiment analysis (ABSA).} 
\end{table*}

\subsection{Aspect-Based Sentiment Analysis (ABSA)}

Aspect-based sentiment analysis (ABSA) extends sentiment analysis by introducing a finer-grained level of task granularity. Unlike sentiment classification, where the output is typically a binary or multi-class label, ABSA delves deeper, focusing on aspects within the sentiment analysis process.

\subsubsection{General ABSA}

Similar to MAST, general ABSA encompasses various sub-tasks, ranging from simpler single ABSA tasks like aspect term extraction to more complex compound ABSA tasks such as aspect sentiment triplet extraction \cite{Zhang2022ASO}. \cite{Zhang2022ASO} provide a comprehensive survey on general ABSA, we recommend referring to their work for further details on general trends in this direction.

\subsubsection{Arabic ABSA}

Research on Arabic ABSA lags behind that of general ABSA. The majority of existing works in Arabic ABSA primarily address aspect sentiment classification, which essentially translates to sentiment classification applied at the aspect level. Additionally, many studies rely solely on feature-based approaches and traditional machine learning methods. This section will focus exclusively on deep learning-based methods for Arabic ABSA. 

The development of Arabic ABSA parallels that of Arabic sentiment classification, as discussed in Section \ref{section: arabic sentiment classification}. A growing number of studies are employing deep learning and pre-trained language model-based methods. \cite{AlSmadi2017DeepRN} and \cite{Alshammari2020AspectbasedSA} compare traditional machine learning and deep learning methods for Arabic ABSA. \cite{AlDabet2021EnhancingAA} propose different network architectures tailored to specific Arabic ABSA tasks. \cite{Abdelgwad2021ArabicAS} develops a BERT-based model for Arabic ABSA. Table \ref{Table: arabic absa} provides detailed descriptions of methods for Arabic ABSA. 
\subsection{Comparison within Arabic Sentiment Analysis Methods}
Although task-specific methods are dedicated to ASA tasks, the most effective methods are those that combine pre-trained language models and various optimisation techniques. hULMonA \cite{eljundi-etal-2019-hulmona} first demonstrate the effectiveness of pre-trained language models by developing a pre-trained LM for Arabic and fine-tuning it for ASA, which significantly improves the performance. Latter pre-trained Arabic LMs such as AraBERT \cite{antoun-etal-2020-arabert} and ARBERT \cite{abdul-mageed-etal-2021-arbert} further push the boundary of ASA by using Arabic specific tokenisation and pre-training models on dialectal Arabic.
\par Another research direction extends the existing methods using various approaches such as domain adaptation \cite{el-mekki-etal-2021-domain} and data augmentation \cite{Refai2022DataAU}. However, the models based on pre-trained LMs are not computationally efficient, and involve a significant computational overhead, whereas approaches such as \cite{alyafeai-ahmad-2021-arabic} maintain the balance between performance and efficiency by distillation and quantisation. 

\subsection{Gaps and challenges in Arabic Sentiment Analysis}\label{sec:gaps}
\subsubsection{Gaps}
This section outlines key research gaps between Arabic sentiment analysis and general sentiment analysis across three dimensions: 
\begin{itemize}
    \item \textbf{Modality:} Multimodality has recently garnered significant interest within general sentiment analysis, with a surge in research activity \cite{Lai2023MultimodalSA}. However, investigations into multi-modal Arabic sentiment analysis remain limited. Most datasets for Arabic sentiment analysis are restricted to the text modality. 
    \item \textbf{Granularity:} The majority of research in Arabic sentiment analysis focuses solely on Arabic sentiment classification. As evidenced in the previous sections, even studies exploring Arabic MAST and Arabic ABSA often target simpler tasks. 
    Consequently, Arabic sentiment analysis lags behind general sentiment analysis in terms of MAST and ABSA tasks. 
    \item \textbf{Context:} While datasets for general sentiment analysis encompass various levels ranging from document level to aspect level, most datasets for Arabic sentiment analysis remain at the document level. Even some recently released datasets lack annotations at sentence level and aspect level. 
\end{itemize}

\subsubsection{Challenges}
\par The Arabic language is characterized by its high morphological complexity, which entails intricate word formation processes that may obscure meaning \cite{habash2010introduction}. Additionally, the high degree of ambiguity and polysemy inherent in Arabic lexicon complicates semantic interpretation. The presence of negation and the extensive range of dialects further exacerbate these challenges, as they introduce variations that must be meticulously accounted for in linguistic models \cite{elbeltagy2013open}.
\par  Data scarcity and cultural contextualization present additional challenges for Arabic. There is a scarcity of large, labeled datasets for many dialects, making it difficult to train robust models. Moreover, sentiment expression can vary significantly based on cultural nuances, requiring models to understand context beyond mere text.  
\section{Recent trends in Arabic Sentiment Analysis}
Several research efforts are ongoing to develop robust Arabic-specific methods and overcome the challenges presented in Section \ref{sec:gaps}. We organise and present these efforts below:
\subsection{Addressing Dialectal Variations}
The issue of dialectal variation has received significant attention in both task-specific and pre-trained language model-based approaches. \cite{baly-etal-2017-characterization} conducted a characterization study analyzing tweets from different Arab regions, highlighting the importance of addressing the dialectal problems in Arabic SA. The efforts that tackle this challenge are presented below, grouped into the broad approach employed:
\subsubsection{Dataset Creation, Domain Adaptation and Data Augmentation}
\cite{medhaffar-etal-2017-sentiment} addressed this challenge by annotating a corpus specifically for the Tunisian dialect and evaluating their models on data from various dialects. A similar approach to this method was the one proposed by \cite{Guellil2018SentiALGAC}. They presented a method for automatically constructing an Algerian dialect corpus. \cite{el-mekki-etal-2021-domain} introduced an unsupervised domain adaptation method for cross-domain and cross-dialect sentiment analysis in Arabic. \cite{Refai2022DataAU} proposed a data augmentation method specifically designed for Arabic classification tasks using transformer-based models.
\subsubsection{Increasing use of Deep learning} 
Researchers are increasingly using deep learning models, particularly transformer based models, to effectively capture the nuances of different Arabic dialects. For example, ARBERT and MARBERT \cite{abdul-mageed-etal-2021-arbert} were specifically pre-trained on dialectal Arabic to address these dialectal variations. 
\subsubsection{Transfer Learning and Multilingual Models}
Transfer learning approaches are being used to leverage knowledge from models trained on larger datasets in other languages or MSA, facilitating better performance on dialect data with limited resources. Multilingual transformer models like mBERT are also applied for handling multiple Arabic dialects \cite{devlin-etal-2019-bert}.




\subsection{Arabic-specific Tokenization}
Recent research has also explored the importance of developing tokenization methods specifically for the Arabic language. \cite{Alyafeai2021EvaluatingVT} compared the performance of different tokenizers for various Arabic classification tasks. \cite{alkaoud-syed-2020-importance} proposed tokenization strategies specifically tailored for both static and contextual Arabic word embeddings, demonstrating significant performance improvements.The efforts in this direction can be grouped into the following trends:

\textbf{Morphological Analysis} Implementation of advanced morphological analysis tools to accurately identify roots, prefixes, and suffixes, ensuring proper tokenization of complex words. Noteworthy contributions in this area include the MADAMIRA tool, which provides robust morphological analysis and disambiguation for modern written Arabic, showcasing significant improvement in processing complex Arabic morphological structures \cite{pasha2014madamira}. 

\textbf{Dialect-Specific Tokenizers:} Development of tokenization models tailored to specific Arabic dialects to handle regional vocabulary and expressions effectively. The CALIMA-Star project exemplifies efforts to create comprehensive morphological lexicons specific to different Arabic dialects, allowing more precise tokenization and analysis for dialectal data \cite{taji-etal-2018-arabic}. 

\textbf{Contextual Tokenization:} Use of context-aware tokenization methods to understand the meaning of words in context, assisting in disambiguating similar words. Contextual models like AraBERT and its advancements in tokenization strategies demonstrate the power of context-aware embeddings to capture nuanced language variations in sentiment analysis \cite{antoun-etal-2020-arabert}. 

\textbf{Incorporating Diacritics:} Desining tokenizers that handle diacritics appropriately, either by retaining them for analysis or by normalizing words without diacritics while preserving meaning. Research by \cite{alqahtani-etal-2020-multitask} highlights the role of diacritics in enhancing sentiment analysis, emphasizing the necessity for tokenizers that efficiently process diacritized text data without losing critical semantic information. 
\par While these trends have demonstrated improved performance for ASA, significant research efforts need to be directed in order to bridge the gap between ASA and general SA for high-resource languages.


\section{Future Directions}
To conclude, we present promising research directions to foster the development of robust models for Arabic sentiment analysis. \\
\textbf{Creation of richer datasets:} Future efforts should prioritize the development of datasets that encompass richer annotations across the following dimensions: \vspace{-0.3cm}
\begin{itemize}
    \item \textbf{Multimodality:} Datasets that integrate various modalities (text, audio, video) to capture richer sentiment information.
    \vspace{-0.1cm}
    \item \textbf{Fine-grained tasks:} Datasets designed for exploring more intricate sentiment analysis tasks beyond sentiment classification.
    \vspace{-0.1cm}
    \item \textbf{Multi-context annotations:} Datasets with annotations at finer levels (sentence level, aspect level) to facilitate in-depth analysis. 
    \vspace{-0.1cm}
\end{itemize}
\textbf{Multimodal Sentiment Analysis:} While limited research has been conducted in multimodal ASA, leveraging information from multiple modalities holds significant potential for accurate sentiment analysis. Future research should explore effective techniques for incorporating multimodal data and develop robust models for this task. 

\textbf{Large Language Models (LLMs) for ASA:} Recent advancements in LLMs have yielded remarkable performance on various tasks. Arabic LLMs like AceGPT-LL \cite{Huang2023AceGPTLL} have also emerged. However, a systematic analysis of LLMs for sentiment analysis, particularly in the context of Arabic, is lacking. Future research should investigate how to best utilize LLMs for Arabic sentiment analysis. 

\textbf{Interpretable Sentiment Analysis:} Existing Arabic sentiment analysis methods primarily provide final sentiment labels without explanations for their outputs. 
Recent work on improving the interpretability of question answering by examining model reasoning \cite{huang-chang-2023-towards} suggests a promising approach that can be adapted to sentiment analysis. For example, models could be designed to output reasoning steps leading to their final sentiment polarity predictions. 

\textbf{Exploration of Fine-Grained Tasks:} General sentiment analysis research has shifted towards increasingly fine-grained tasks. However, most Arabic sentiment analysis studies remain focused on sentiment classification at the document level. A systematic exploration of other fine-grained tasks, particularly those within MAST and ABSA, would be beneficial for advancing the field. 

\section{Limitations}

This survey acknowledges some limitations. Firstly, it primarily focuses on works utilizing deep learning methods. As highlighted in \cite{abu-kwaik-etal-2022-pre}, feature-based methods can outperform pre-trained language model-based methods in some instances. Future surveys may benefit from including an exploration of feature-based approaches. Additionally, while this work compares Arabic sentiment analysis with general sentiment analysis, it would also be valuable to situate Arabic sentiment analysis within the broader context of Arabic classification tasks and Arabic natural language processing tasks in general. 
\bibliography{acl_latex}

\end{document}